\documentclass{article}

\usepackage[preprint]{neurips_2022}

\usepackage[dvipsnames]{xcolor}         
\definecolor{linkColor}{rgb}{0.18,0.39,0.62}
\usepackage[utf8]{inputenc} 
\usepackage[T1]{fontenc}    
\usepackage[colorlinks=true,linkcolor=linkColor,citecolor=linkColor,filecolor=linkColor,urlcolor=linkColor]{hyperref}       
\usepackage{url}            
\usepackage{booktabs}       
\usepackage{amsfonts}       
\usepackage{nicefrac}       
\usepackage{microtype}      

\usepackage{amsmath}
\usepackage{amssymb}
\usepackage{mathtools}
\usepackage{amsthm}
\usepackage{graphicx}
\usepackage{subfigure}
\usepackage{xspace}
\usepackage{pifont}

\theoremstyle{plain}

\theoremstyle{definition}

\theoremstyle{remark}

\usepackage[capitalize,noabbrev]{cleveref}
\usepackage{framed}
\usepackage{multirow}
\usepackage{svg}
\usepackage{listings}
\usepackage{wrapfig}

\newcommand\our{\textsc{TorchScale}}
\newcommand\deepnet{DeepNet}
\newcommand\magneto{Magneto}

\title{\our{}: Transformers at Scale}

\author{
\textbf{Shuming Ma\thanks{~Equal contribution. $\dagger$ Corresponding author.},~\ Hongyu Wang\footnotemark[1],~\ Shaohan Huang,~Wenhui Wang,~Zewen Chi,~Li Dong} \\ ~\textbf{Alon Benhaim,}~\textbf{Barun Patra,}~\textbf{Vishrav Chaudhary,}~\textbf{Xia Song,}~\textbf{Furu Wei}$^\dagger$ \\
Microsoft\\
{\url{https://aka.ms/torchscale}}
}

\begin{document}

\maketitle

\begin{abstract}
Large Transformers have achieved state-of-the-art performance across many tasks. Most open-source libraries on scaling Transformers focus on improving training or inference with better parallelization. In this work, we present \our{}, an open-source toolkit that allows researchers and developers to scale up Transformers efficiently and effectively. \our{} has the implementation of several modeling techniques, which can improve modeling generality and capability, as well as training stability and efficiency. Experimental results on language modeling and neural machine translation demonstrate that \our{} can successfully scale Transformers to different sizes without tears. The library is available at \url{https://aka.ms/torchscale}.
\end{abstract}

\section{\our{}: A Library for Transformers at (Any) Scale}

Recently, there is a trend that the Transformers have become the de facto backbone across different areas, and their model sizes are growing large.
Scaling Transformers brings more capacity and the convergence is much faster compared to the smaller models. More importantly, more emergent abilities are observed in the larger models, and they are unpredictable in the smaller models~\citep{emergent}. Therefore, both researchers and practitioners should benefit from a toolkit that can scale Transformers at any scale. While there are some toolkits on scaling Transformers, most of them focus on improving the parallelization of the systems, such as model parallelism~\citep{megatron}, pipeline parallelism~\citep{gpipe}, expert parallelism~\citep{gshard}, and optimizer parallelism~\citep{zero}. Different from them, \our{} improves the scalability of Transformers from the modeling perspective. It implements several modeling techniques that can improve generality, stability, and efficiency during scaling up the model size. 

\subsection{Generality}

There are multiple Transformer~\citep{transformer} variants for different tasks, including Post-LN, Pre-LN, and so on.
\our{} adopts \magneto{}~\citep{magneto} as the default backbone. Magneto is a foundation Transformer for general-purpose modeling, which serves as a go-to architecture for various tasks and modalities. With one unified architecture, this allows \our{} to support different applications, including language modeling, machine translation, vision pretraining, speech recognition, and multi-modal pretraining. \magneto{} is simple and efficient. Compared to Pre-LN, \magneto{} introduces an extra LayerNorm into each sublayer. We refer more details to~\citet{magneto}. \our{} has the implementation of encoder, decoder, and encoder-decoder architectures for different tasks.

\subsection{Stability}

As the model grows, one problem is that Transformers are more unstable in optimization. The models can diverge at any time during the training. This requires lots of computation and human efforts to tune and test the models. To get rid of the pain when scaling up the models, \our{} follows the theoretical derivation of \deepnet{}~\citep{deepnet} to improve the training stability. Specifically, we implement the initialization method of Sub-LN for \magneto{}. Besides, as an alternative, we also implement the DeepNorm from DeepNet, which can fundamentally improve the optimization of Post-LN Transformers. With better stability, the models accept larger learning rate as well as more diverse data that may contain some noise. Moreover, it enables some modeling techniques that might bring stability issues.

\subsection{Efficiency}

One mainstream of scaling Transformers is the mixture-of-experts (MoE) model. MoE models introduce the sparsity to Transformers, which proves to be both effective and efficient. \our{} implement X-MoE~\citep{xmoe}, a variant of sparse MoE model. It leverages the sparsity of mixture-of-experts while preserving the transferability of Transformers by mitigating the representation collapse problem of the sparse MoE models. \our{} supports both Top-1 and Top-2 routing algorithms, which balance the performance and the computation cost. This allows Transformers to scale up to billions or trillions of parameters without much additional computation cost. In our preliminary experiments, we observe that gradient clipping plays an important role in the performance of sparse MoE models. We propose a novel gradient clipping method, called SparseClip, which improves the training stability while preserving the performance of the sparse models.

\section{Scaling Up Experiments}

We conduct experiments on language modeling and neural machine translation. With \our{}, we train the models with the architectures of both decoder and encoder-decoder. We further scale up their depths and widths to evaluate the stability and convergence.

\begin{figure}[t]
\centering
\subfigure[Scaling Depth]{
\includegraphics[width=0.45\columnwidth]{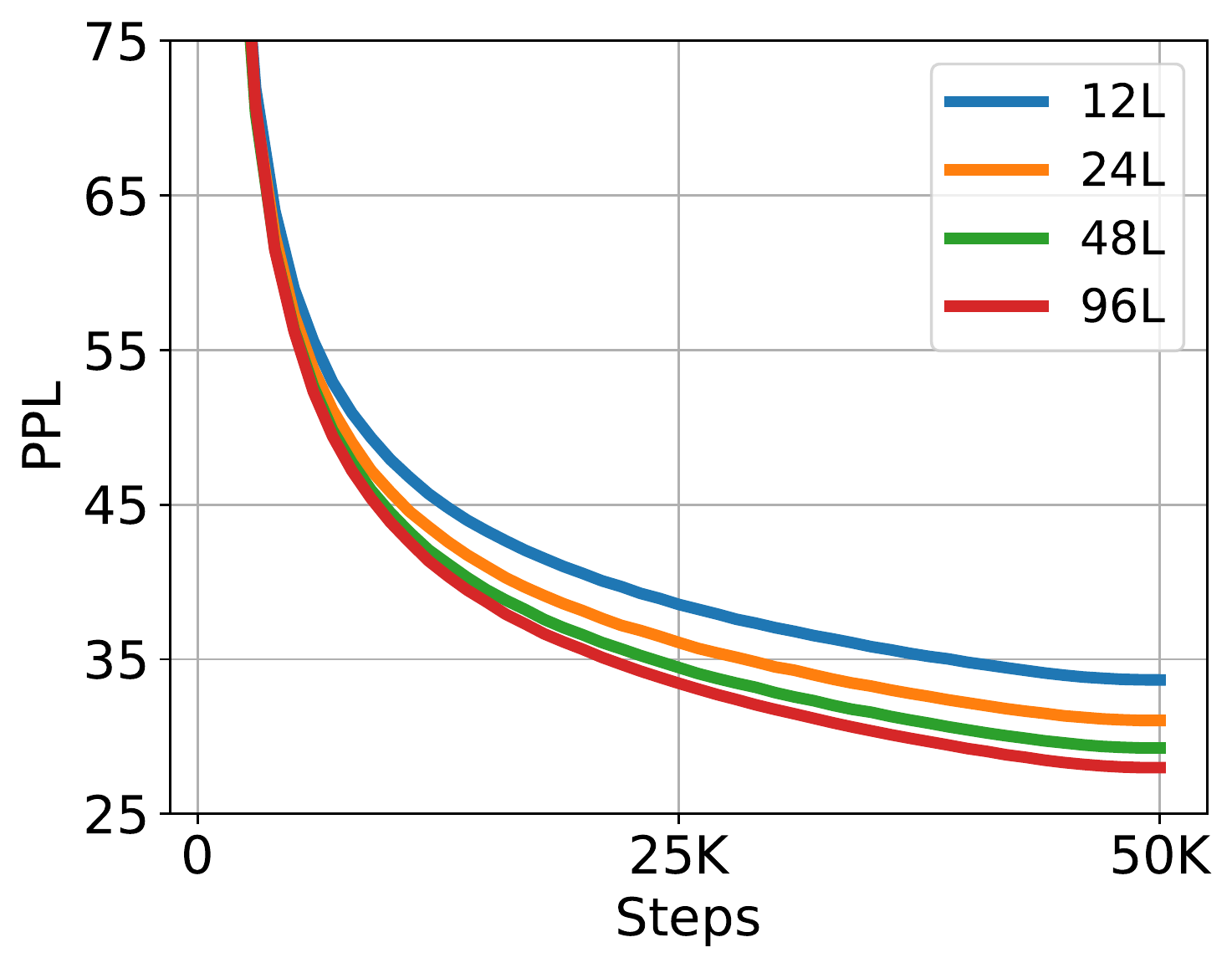}\label{fig:lm_dense}
}
\subfigure[Scaling Width]{
\includegraphics[width=0.45\columnwidth]{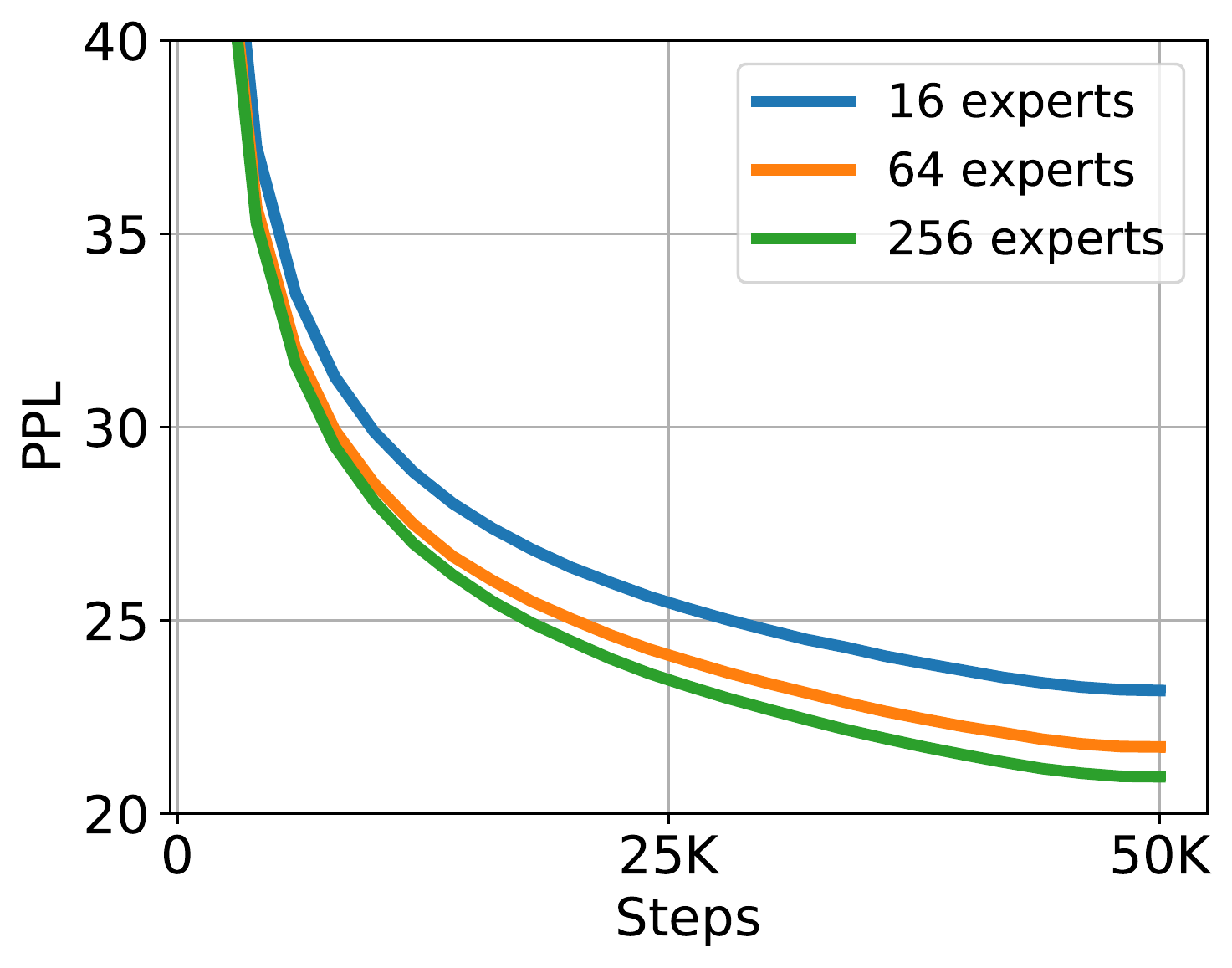}\label{fig:lm_sparse}
}
\caption{Scaling-up experiments on language modeling.
}
\label{fig:lm}
\end{figure}

%
%
\begin{figure}[t]
\centering
\subfigure[Scaling Depth]{
\includegraphics[width=0.45\columnwidth]{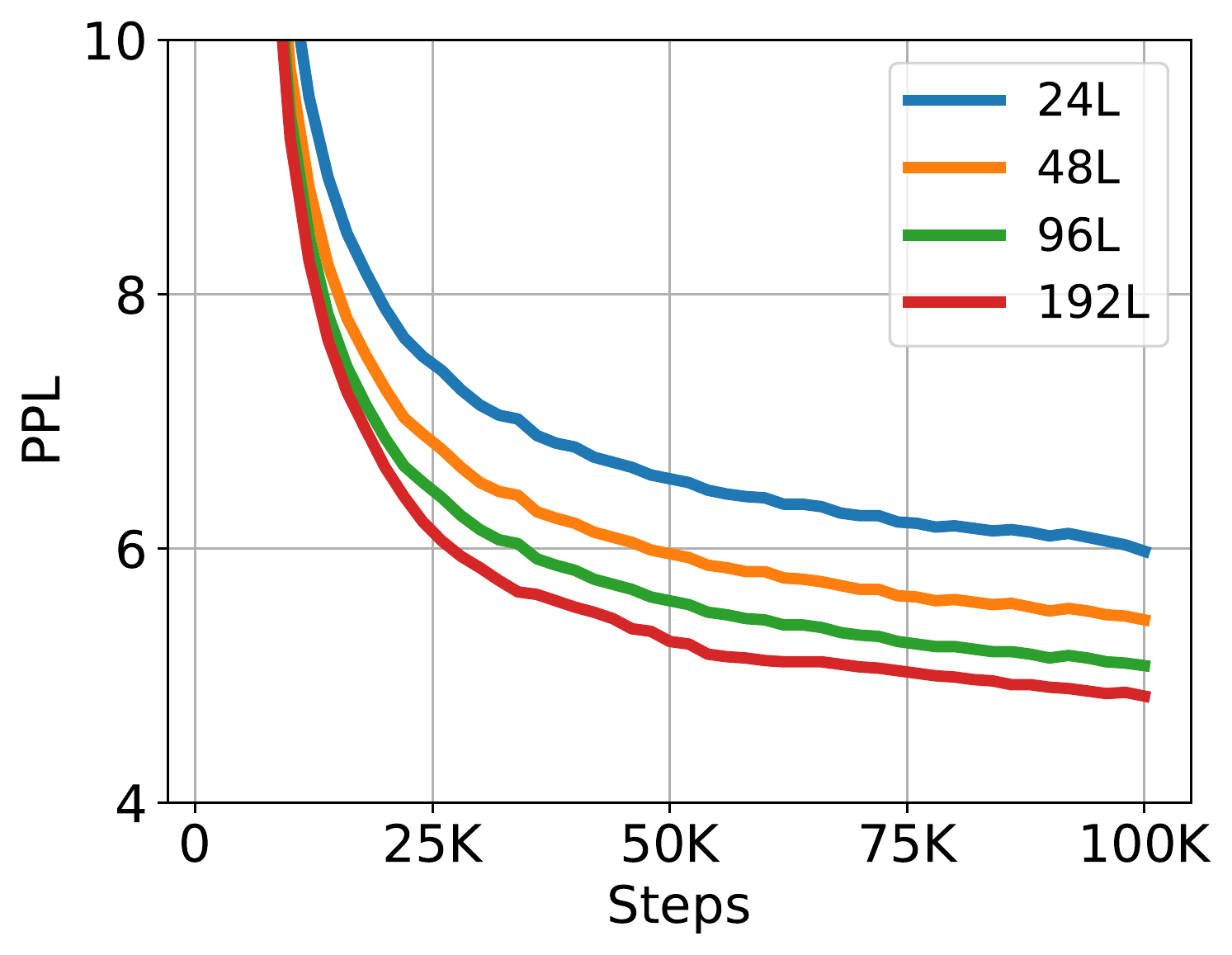}\label{fig:mt_dense}
}
\subfigure[Scaling Width]{
\includegraphics[width=0.45\columnwidth]{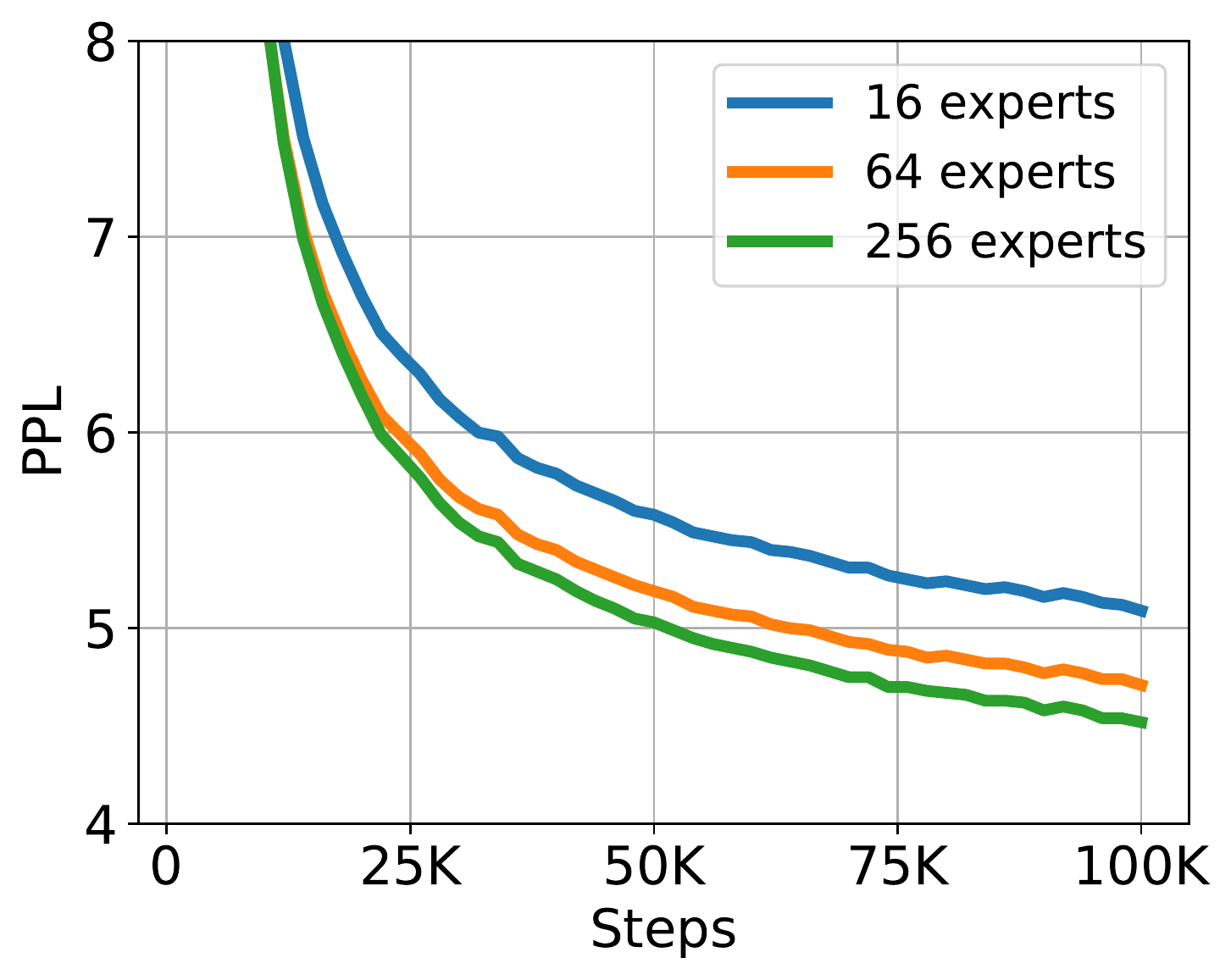}\label{fig:mt_sparse}
}
\caption{Scaling-up experiments on neural machine translation.
}
\label{fig:mt}
\end{figure}

\subsection{Language Modeling}

The experiments on language modeling are performed on an English-language corpus, which is a subset of the data from~\citet{roberta} and the English portion of CC100 corpus~\citep{xlmr}. The validation data is randomly sampled from the same corpus. We preprocess the data with the GPT-2~\citep{gpt-2} tokenizer. We train the models with a batch size of 2048 samples. Each sample contains 128 tokens. The models are trained for 50 thousand steps with up to 13 billion tokens in total. More training details are in the appendix.

We start from a \magneto{} model with 12 decoder layers, 1024 hidden size, and 16 attention heads, and further scale its depth from 12L to 24L, 48L, and 96L. We evaluate their perplexity (PPL) on the validation set and plot the curves in~\cref{fig:lm_dense}. It shows that \our{} can successfully scale up the depth. With better training stability, the convergence is smooth regardless of the depth. Moreover, the expressive power grows as the models deepen, resulting in better validation PPL. 

In addition to the depth, we also scale the width. To scale it up efficiently, we replace the feed-forward layers with the X-MoE layers and gradually increase the number of experts from 16 experts to 64experts and 256 experts. We use top-2 routing as it produces better results in our preliminary experiments. \cref{fig:lm_sparse} summarizes the results of the sparse MoE models. Although a larger model size brings more challenges in optimization, \our{} can still successfully train a model with 256 experts and up to 13B parameters, outperforming the models with a smaller size. This proves the effectiveness of \our{} to train the Transformers at any scale.

\subsection{Neural Machine Translation}

We evaluate \our{} on neural machine translation. The experiments are conducted using a combination of CCMatrix~\citep{ccmatrix}, CCAligned~\citep{ccaligned}, OPUS~\citep{opus100}, and Tatoeba\footnote{\url{https://tatoeba.org/en/}}. The final data consists of 102 languages, 1932 directions, and 12B sentence pairs. We tokenize the data with the SentencePiece model provided by the Flores-101 dataset~\citep{flores101}. We use a batch size of 262K tokens to train the model for 100K steps. More details can be found in the appendix.

With \our{}, we implement a \magneto{} model in the architecture of encoder-decoder. The model has a 12-layer encoder and a 12-layer decoder with 1024 hidden dimension, 16 heads, and 4096 intermediate dimension of feed-forward layers. Similar to the experiments on language modeling, we scale the number of layers from 12L-12L to 24L-24L, 48L-48L, and 96L-96L. \cref{fig:mt_dense} shows that the training of these models is stable. More importantly, better performance can be achieved as the model grows. This proves that \our{} can fully use the capacity of a larger model.

We also evaluate the sparse X-MoE models with top-2 routing. \cref{fig:mt_sparse} illustrates the scaling curves from 16 experts, 64 experts, to 256 experts. \our{} can effectively scale up the number of experts, reaching a significant improvement over the smaller models. Notably, the training flops are almost the same among these models, proving the efficiency of \our{}. 

\subsection{SparseClip: Gradient Clipping for Sparse MoE Models}

Gradient clipping is standard practice for deep neural models to alleviate the gradient explosion problem. It is even more important for the sparse MoE models, which are more unstable in training. Vanilla gradient clipping can be formulated as:
\begin{equation}
    g \leftarrow \frac{\xi}{\max(\xi, ||g||_2)}\ g,
\end{equation}
where $\xi$ is a hyper-parameter to control the strength of clipping. It scales the model's gradient $g$ when its L2 norm $||g||_2$ exceeds $\xi$. For the sparse MoE models, the gradient norm is computed as:
\begin{equation}
    ||g||_2 = ||\ (g_{d},\ \ g_{s})\ ||_2,
\end{equation}
where $g_{d}$ is the gradients of the dense part and $g_s$ is the gradients of the MoE parameters.

\begin{figure}[t]
\centering
\includegraphics[width=0.45\columnwidth]{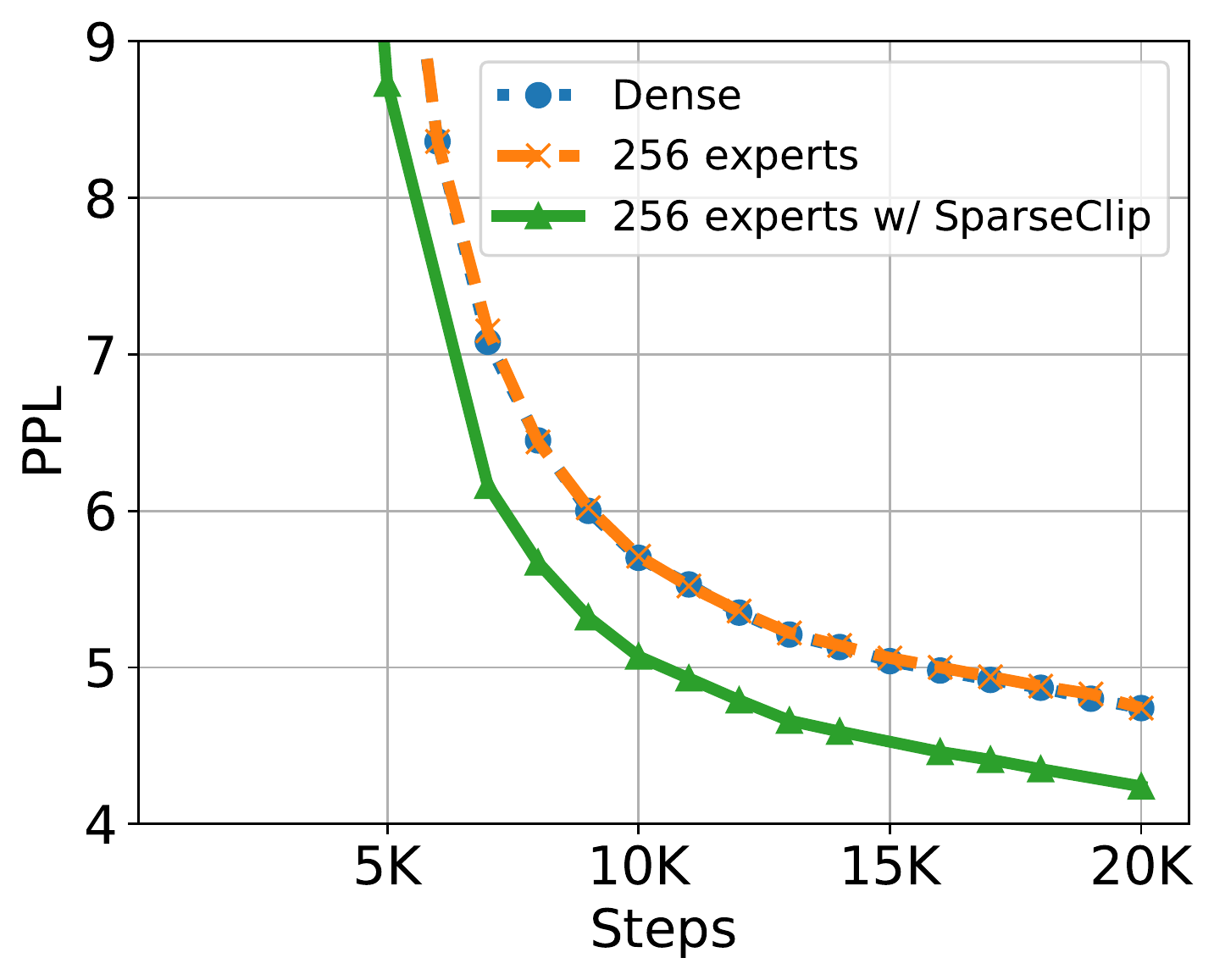}
\caption{Training curves on neural machine translation. SparseClip significantly improves the convergence of the sparse MoE models.
}
\label{fig:clip}
\end{figure}

It is natural to directly combine the gradients of two parts before computing the L2 norm of the whole model.
However, when the MoE models grow, the MoE gradients become to dominate the gradient norm, making the gradient direction inaccurate after clipping the gradients. We conduct experiments to compare a 256-expert model with its dense counterpart, which both use the vanilla gradient clipping during training. \cref{fig:clip} shows that the convergence of the 256-expert model has no significant difference with the dense model, despite the fact that it has over 100x parameters than the dense model. It proves that the vanilla gradient clipping hurts the sparse MoE models, limiting their scaling to larger model sizes. Meanwhile, removing gradient clipping is infeasible as it will destabilize the model training and results in divergence at the beginning of training.

To deal with this problem, we propose SparseClip, which re-weights the gradients of MoE parameters when computing the gradient norm. Specifically, the gradient norm can be calculated as:
\begin{equation}
    ||g||_2 = ||\ (g_{d},\ \ \kappa g_{s})\ ||_2,
\end{equation}
\begin{equation}
    \kappa = \frac{1}{\sqrt{E}},
\end{equation}
where $E$ is the number of experts for sparse MoE models. This can mitigate the domination of the MoE gradients, making the total gradient norm nearly constant with the increase of experts.

\cref{fig:clip} compares SparseClip with vanilla gradient clipping. It shows that Sparseclip significantly outperforms the vanilla gradient clipping, leading to a lower PPL. This verifies the effectiveness of our SparseClip method.

\begin{figure}[t]
\centering
\subfigure[Training Entropy of Routing]{
\includegraphics[width=0.45\columnwidth]{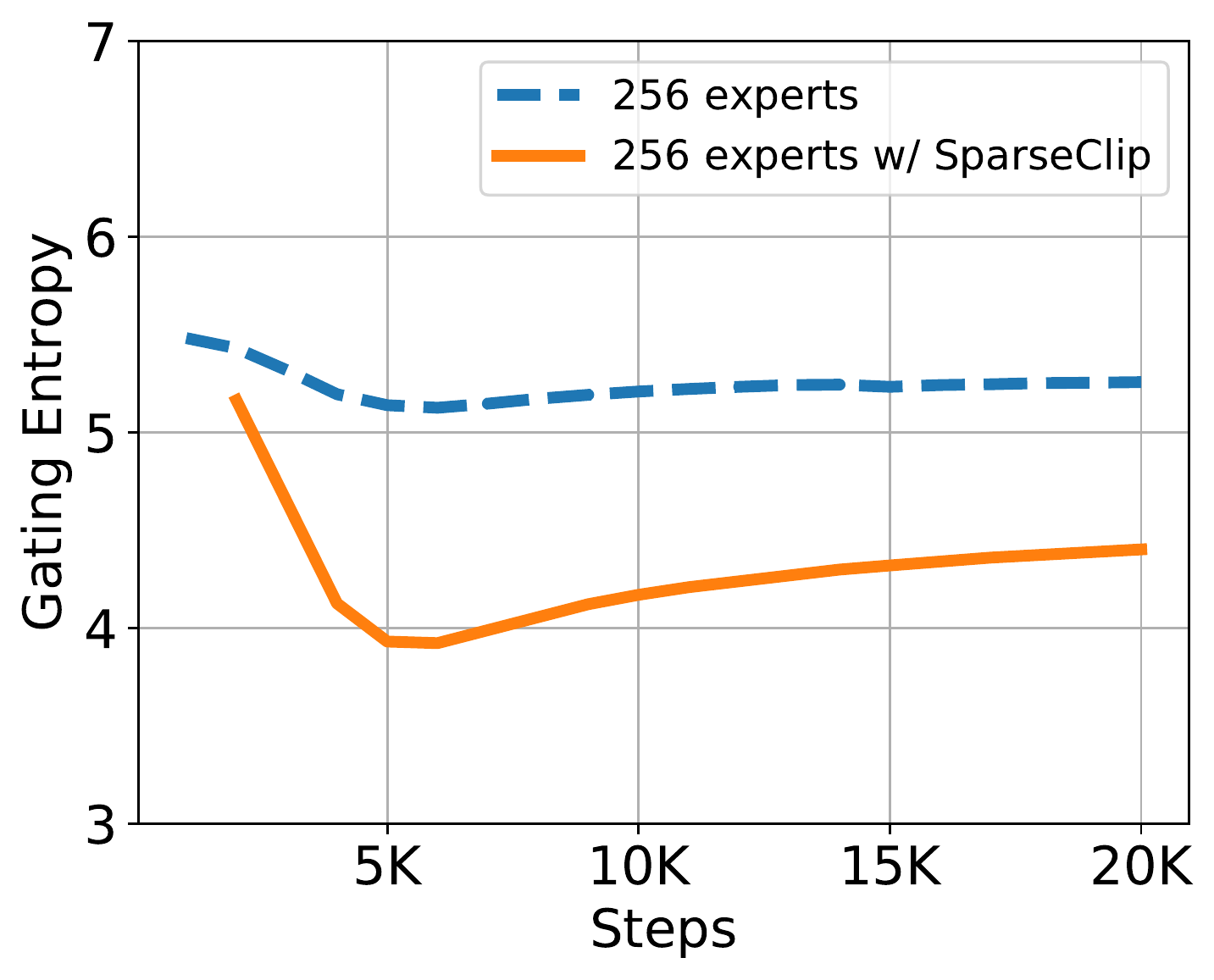}\label{fig:gate}
}
\subfigure[Number of Gradient Clipping Triggered]{
\includegraphics[width=0.45\columnwidth]{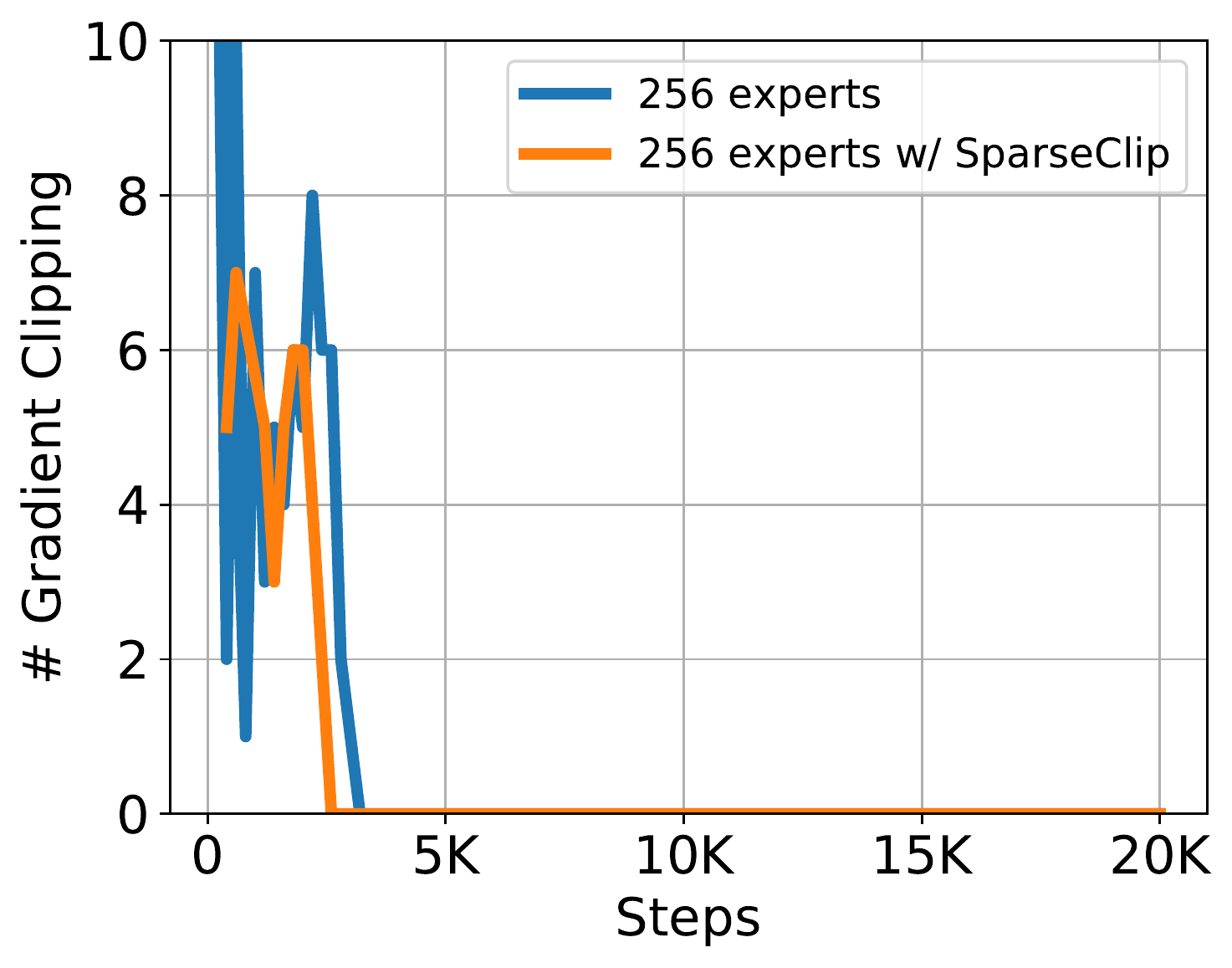}\label{fig:norm}
}
\caption{Visualization of SparseClip and vanilla gradient clipping on neural machine translation.
}
\label{fig:sparseclip}
\end{figure}

We further perform some visualizations of SparseClip and the vanilla gradient clipping. \cref{fig:gate} is the training entropy of routing. It shows that SparseClip has lower entropy than the vanilla gradient clipping, indicating that SparseClip improves the routing of MoE layers. \cref{fig:norm} illustrates the number of gradient clippings. While gradient clipping is triggered only at the beginning of the training, it is interesting that it has a great effect on the performance throughout the training. We leave further research on it in the future work.

\section{Conclusion}

We introduce \our{}, an open-source toolkit that enables scaling Transformers both efficiently and effectively. It implements several state-of-the-art modeling techniques, originally from Magneto, DeepNet, and X-MoE, to improve modeling generality and capability, as well as training stability and efficiency. \our{} adopts a novel gradient clipping method, called SparseClip, which can further improve the stability while preserving the performance of large sparse models. The evaluations on language modeling and machine translation demonstrate that \our{} allows scaling Transformers 
at different model sizes.

\bibliographystyle{plainnat}
\bibliography{torchscale}

\newpage

\appendix

\section{Hyperparameters}

\begin{table}[ht]
\centering
\begin{tabular}{l|c}
\toprule
\bf Hyperparameters & \bf Language Modeling \\
\midrule
Layers & \{12, 24, 48, 96\} \\
Hidden size & 1024 \\
FFN inner hidden size & 4096 \\
Attention heads & 16 \\
\midrule
Peak learning rate & 5e-4 \\
Batch size & 2048 \\
Adam $\beta$ & (0.9, 0.98) \\
Learning rate schedule & Polynomial decay \\
Warmup updates & 750 \\
\midrule
Dropout & 0.1 \\
Attention dropout & 0.1 \\
Weight decay & 0.01 \\
\bottomrule
\end{tabular}
\caption{
Hyperparameters of dense models for the experiments on language modeling. 
}
\label{tbl:hyper:lm:dense}
\end{table}

\begin{table}[ht]
\centering
\begin{tabular}{l|c}
\toprule
\bf Hyperparameters & \bf Language Modeling \\
\midrule
Layers & 12 \\
Experts & \{16, 64, 256\} \\
Hidden size & 1024 \\
FFN inner hidden size & 4096 \\
Attention heads & 16 \\
MoE Frequency & 2 \\
Weight of Balance Loss & 0.01 \\
\midrule
Peak learning rate & 5e-4 \\
Batch size & 2048 \\
Adam $\beta$ & (0.9, 0.98) \\
Learning rate schedule & Polynomial decay \\
Warmup updates & 750 \\
\midrule
Dropout & 0.1 \\
Attention dropout & 0.1 \\
Weight decay & 0.01 \\
\bottomrule
\end{tabular}
\caption{
Hyperparameters of sparse models for the experiments on language modeling. 
}
\label{tbl:hyper:lm:sparse}
\end{table}

\begin{table}[ht]
\centering
\begin{tabular}{l|c}
\toprule
\bf Hyperparameters & \bf Neural Machine Translation \\
\midrule
Layers & \{12-12, 24-24, 48-48, 96-96\} \\
Hidden size & 1024 \\
FFN inner hidden size & 4096 \\
Attention heads & 16 \\
\midrule
Peak Learning rate & 5e-4 \\
Learning rate schedule & Inverse sqrt \\
Warm-up updates & 6000 \\
Batch Size (tokens) & 262K \\
Adam $\beta$ & (0.9, 0.98) \\
Label smoothing & 0.1 \\
\midrule
Gradient clipping & 1.0 \\
Dropout & 0.1 \\
Weight decay & 0.0 \\
\bottomrule
\end{tabular}
\caption{
Hyperparameters for dense models for the experiments on neural machine translation. 
}
\label{tbl:hyper:mt:dense}
\end{table}

\begin{table}[ht]
\centering
\begin{tabular}{l|c}
\toprule
\bf Hyperparameters & \bf Neural Machine Translation \\
\midrule
Layers & 12-12 \\
Experts & \{16, 64, 256\} \\
Hidden size & 1024 \\
FFN inner hidden size & 4096 \\
Attention heads & 16 \\
MoE Frequency & 2 \\
Weight of Balance Loss & 0.01 \\
\midrule
Peak Learning rate & 5e-4 \\
Learning rate schedule & Inverse sqrt \\
Warm-up updates & 6000 \\
Batch Size (tokens) & 262K \\
Adam $\beta$ & (0.9, 0.98) \\
Label smoothing & 0.1 \\
\midrule
Gradient clipping & 1.0 \\
Dropout & 0.1 \\
Weight decay & 0.0 \\
\bottomrule
\end{tabular}
\caption{
Hyperparameters for sparse models for the experiments on neural machine translation. 
}
\label{tbl:hyper:mt:sparse}
\end{table}

\end{document}